\documentclass[article]{siamart190516}
\usepackage{amsmath,amssymb, mathdots}
\usepackage{caption}
\usepackage{hyperref}
\usepackage[utf8]{inputenc}
\usepackage{mathtools}
\usepackage{color,graphicx}
\usepackage{thmtools, thm-restate}
\usepackage{enumitem}
\usepackage{stmaryrd}
\usepackage{tensor}
\newcommand*{\rom}[1]{\expandafter\@slowromancap\romannumeral #1}
\reversemarginpar

\usepackage{algorithm}
\usepackage{algpseudocode}

\usepackage{tikz}
\usetikzlibrary{patterns}
\usetikzlibrary{arrows}
\usetikzlibrary{3d,calc}

\usepackage[colorinlistoftodos,textsize=tiny,textwidth=31mm]{todonotes}
\usepackage{etoolbox}
\usepackage{mathabx}

\makeatletter
\newcommand*\bigcdot{\mathpalette\bigcdot@{.5}}
\newcommand*\bigcdot@[2]{\mathbin{\vcenter{\hbox{\scalebox{#2}{$\m@th#1\bullet$}}}}}
\makeatother

\makeatletter

\patchcmd{\@addmarginpar}{\ifodd\c@page}{\ifodd\c@page\@tempcnta\m@ne}{}{}
\providecommand*{\cupdot}{%
  \mathbin{%
    \mathpalette\@cupdot{}%
  }%
}
\newcommand*{\@cupdot}[2]{%
  \ooalign{%
    $\m@th#1\cup$\cr
    \hidewidth$\m@th#1\cdot$\hidewidth
  }%
}
\providecommand*{\bigcupdot}{%
  \mathop{%
    \vphantom{\bigcup}%
    \mathpalette\@bigcupdot{}%
  }%
}
\newcommand*{\@bigcupdot}[2]{%
  \ooalign{%
    $\m@th#1\bigcup$\cr
    \sbox0{$#1\bigcup$}%
    \dimen@=\ht0 %
    \advance\dimen@ by -\dp0 %
    \sbox0{\scalebox{2}{$\m@th#1\cdot$}}%
    \advance\dimen@ by -\ht0 %
    \dimen@=.5\dimen@
    \hidewidth\raise\dimen@\box0\hidewidth
  }%
}

\makeatother
\reversemarginpar
\newsiamremark{expl}{Example}

\newcommand{\cN}{\mathcal{N}}

\newcommand{\cO}{\mathcal{O}}

\newcommand{\cR}{\mathcal{R}}

\newcommand{\cX}{\mathcal{X}}

\newcommand{\bitm}{\begin{itemize}}
\newcommand{\eitm}{\end{itemize}}
\newcommand{\bitme}{\begin{enumerate}[label=(\roman*),leftmargin=0.25in]}
\newcommand{\eitme}{\end{enumerate}}
\newcommand{\beq}{\begin{equation}}
\newcommand{\eeq}{\end{equation}}
\newcommand{\btcb}{\begin{tcolorbox}}
\newcommand{\etcb}{\end{tcolorbox}}
\def\bals#1\eals{\begin{align*} #1 \end{align*}}
\def\bal#1\eal{\begin{align} #1 \end{align}}

\newcommand{\where}{\quad \text{ where } }

\newcommand\Dom\Omega

\newcommand\EE{\mathbb{E}}

\newcommand\RR{\mathbb{R}}

\newcommand\NN{\mathbb{N}}

\newcommand\Lap\Delta

\newcommand\abs[1]{\left\lvert #1 \right\rvert}

\def\bpde#1\epde{\[\left\{\begin{aligned}#1\end{aligned}\right. \]}
\def\inbpde#1\inepde{\left\{\begin{aligned}#1\end{aligned}\right.}
\def\binpde#1\einpde{\left\{\begin{aligned}#1\end{aligned}\right.}

\newcommand\Norm[2]{\lVert { #1 } \rVert_{#2}}

\def\cB{\mathcal{B}}

\def\cD{\mathcal{D}}

\def\cL{\mathcal{L}}

\def\cR{\mathcal{R}}

\def\b0{\mathbf{0}}

\def\bbmat{\begin{bmatrix}[r]}
\def\ebmat{\end{bmatrix}}

\newcommand{\barr}{\begin{array}}
\newcommand{\ea}{\end{array}}
\newcommand{\bea}{\begin{eqnarray}}
\newcommand{\eea}{\end{eqnarray}}
\newcommand{\bt}{\begin{table}}
\newcommand{\et}{\end{table}}

\DeclareMathOperator\diag{diag}

\numberwithin{equation}{section}

\newcommand\tturl[1]{{\tt \scriptsize [\url{{#1}}]}}

\newcommand{\TheTitle}{%
F{\small ast}LRNR and Sparse Physics Informed Backpropagation}

\newcommand{\remove}[1]{\textcolor{red}{[Text removed.]}}%

\graphicspath{{../figures/}{../python/}}

\newcommand{\RN}[1]{%
\textup{\uppercase\expandafter{\romannumeral#1}}%
}

\newcommand{\Mell}[0]{{M_{\ell}}}
\newcommand{\Mhyperell}[0]{{M_{\textrm{hyper},\ell}}}
\newcommand{\Mellm}[0]{{M_{\ell-1}}}
\newcommand{\Mmax}[0]{{M_{\textrm{max}}}}
\newcommand{\Mtotal}[0]{{M_{\textrm{total}}}}
\newcommand{\Mmaxz}[0]{M^2_{\textrm{max}}}

\newcommand{\rell}[0]{{r_{\ell}}}
\newcommand{\rmin}[0]{{r_{\textrm{min}}}}
\newcommand{\rmax}[0]{{r_{\textrm{max}}}}
\newcommand{\rmaxz}[0]{{r^2_{\textrm{max}}}}
\newcommand{\rtotal}[0]{{r_{\textrm{total}}}}

\newcommand{\rellp}[0]{{r_{\ell+1}}}
\newcommand{\hrell}[0]{\hat{r}_{\ell}}

\newcommand{\hrellmax}[0]{\hat{r}_{\textrm{max}}}

\newcommand{\Well}[0]{W^{\ell}}

\newcommand{\bell}[0]{b^{\ell}}

\newcommand{\hUell}[0]{\widehat{U}^{\ell}}

\newcommand{\Ufix}[0]{U_*}
\newcommand{\Vfix}[0]{V_*}
\newcommand{\bfix}[0]{b_*}
\newcommand{\Thetafix}[0]{\Theta_*}
\newcommand{\Uell}[0]{U^{\ell}}
\newcommand{\Uellt}[0]{U^{\ell \top}}

\newcommand{\hVellp}[0]{\widehat{V}^{\ell+1}}
\newcommand{\hVellpt}[0]{\widehat{V}^{(\ell+1) \top}}

\newcommand{\Vell}[0]{V^{\ell}}
\newcommand{\Vellt}[0]{V^{\ell \top}}
\newcommand{\Vellpt}[0]{V^{(\ell+1) \top}}
\newcommand{\Pell}[0]{P^{\ell}}
\newcommand{\Pellt}[0]{P^{\ell \top}}

\newcommand{\Sell}[0]{S^{\ell}}
\newcommand{\sell}[0]{s^{\ell}}
\newcommand{\zinput}[0]{z^{0}}
\newcommand{\zoutput}[0]{z^{L}}
\newcommand{\zell}[0]{z^{\ell}}
\newcommand{\zellm}[0]{z^{\ell - 1}}

\newcommand{\hzell}[0]{\hat{z}^{\ell}}

\newcommand{\hzellm}[0]{\hat{z}^{\ell - 1}}

\newcommand{\zetaell}[0]{\zeta^{\ell}}
\newcommand{\zetaelleim}[0]{\zeta^{\ell}_{\text{\tiny EIM}}}

\newcommand{\Xiell}[0]{\Xi^{\ell}}

\newcommand{\fhyper}[0]{f_{\textrm{hyper}}}
\newcommand{\umeta}[0]{u_{\textrm{\tiny META}}}
\newcommand{\uinr}[0]{u_{\textrm{\tiny INR}}}
\newcommand{\ulrnr}[0]{u_{\textrm{\tiny LRNR}}}
\newcommand{\ufast}[0]{\hat{u}_{\textrm{\tiny FAST}}}
\newcommand{\ufastsol}[0]{u_{\textrm{\tiny FASTSOL}}}

\newcommand{\muimin}[0]{\mu_{\textrm{conv},\textrm{min}}}
\newcommand{\muimax}[0]{\mu_{\textrm{conv},\textrm{max}}}

\newcommand{\muiimin}[0]{\mu_{\textrm{cdr},\textrm{min}}}
\newcommand{\muiimax}[0]{\mu_{\textrm{cdr},\textrm{max}}}

\newcommand{\regparamsparse}[0]{\lambda_{\textrm{sparse}}}
\newcommand{\regparamortho}[0]{\lambda_{\textrm{orth}}}
\newcommand{\regparamlocal}[0]{\lambda_{\textrm{loc}}}

\newcommand{\regsparse}[0]{\cR_{\textrm{sparse}}}
\newcommand{\regortho}[0]{\cR_{\textrm{orth}}}
\newcommand{\reglocal}[0]{\cR_{\textrm{loc}}}

\newcommand{\lossmeta}[0]{\cL_{\textrm{meta}}}
\newcommand{\losstune}[0]{\cL_{\textrm{tune}}}
\newcommand{\lossfast}[0]{\cL_{\textrm{fast}}}

\newcommand{\domone}[0]{\cD_{\textrm{conv}}}
\newcommand{\domtwo}[0]{\cD_{\textrm{cdr}}}

\begin{document}

\ifpdf
  \DeclareGraphicsExtensions{.pdf, .jpg, .tif}
\else
  \DeclareGraphicsExtensions{.eps, .jpg}
\fi

\title{\TheTitle}

\author{%
  Woojin Cho\thanks{{\scriptsize Telepix, Seoul, 07330, South Korea ({\tt woojin@telepix.net})}}%
  \and 
  Kookjin Lee\thanks{{\scriptsize Department of Computer Science, %
  Arizona State University, Phoenix, AZ 85281, USA ({\tt kookjin.lee@asu.edu})}}%
  \and 
  Noseong Park\thanks{{\scriptsize School of Computing, %
  Korea Advanced Institute of Science and Technology, Daejeon, 34141, South Korea ({\tt noseong@kaist.ac.kr})}} %
  \and {}\\
  Donsub Rim\thanks{{\scriptsize %
  Department of Mathematics, %
  Washington University in St. Louis, St. Louis, MO 63130, USA %
  ({\tt rim@wustl.edu})}}%
  \and 
  Gerrit Welper\thanks{{\scriptsize Department of Mathematics, %
  University of Central Florida,
  Orlando, FL 32816, USA ({\tt gerrit.welper@ucf.edu})}}%
}
\maketitle

\begin{abstract}
We introduce \emph{S}parse \emph{P}hysics \emph{In}formed
Back\emph{prop}agation (SPInProp), a new class of methods for accelerating
backpropagation for a specialized neural network architecture called Low Rank
Neural Representation (LRNR). The approach exploits the low rank structure
within LRNR and constructs a reduced neural network approximation that is much
smaller in size.  We call the smaller network FastLRNR.  We show that
backpropagation of FastLRNR can be substituted for that of LRNR, enabling a
significant reduction in complexity. We apply SPInProp to a physics informed
neural networks framework and demonstrate how the solution of parametrized
partial differential equations is accelerated.
\end{abstract}


\begin{AMS}
\texttt{68T07}, \texttt{65D25}, \texttt{65M22}
\end{AMS}

\section{Introduction} \label{sec:intro}

Backpropagation is a key concept used in training deep learning
models~\cite{goodfellow2016}. This paper concerns a technique for accelerating
backpropagation via dimensionality reduction in a specialized neural network
(NN) architecture called Low Rank Neural Representation (LRNR)
\cite{cho2023,rim2024}. LRNRs have a built-in low rank factorization structure  
that grants them advantages in solving parametrized partial differential
equations (pPDEs)~\cite{cho2023}, in particular for problems that proved
challenging for physics informed neural networks (PINNs)
\cite{raissi2019,krishnapriyan2021}. Theoretically, LRNRs are able to
approximate complicated nonlinear shock interactions while maintaining low
dimensionality and regular dynamics \cite{rim2024}.

We show that, thanks to the LRNRs' low rank structure, smaller NN approximations
we call FastLRNRs can be constructed. Backpropagation can be performed on the
FastLRNR efficiently, and the resulting derivatives can be used to approximate
derivatives of the original LRNR.  Here we focus on presenting the main ideas
and computationally study the simplest possible version. We anticipate many
variants exploiting the same structure, and we refer to this class of methods 
accelerating backpropagation as \emph{Sparse Physics Informed Backpropagation}
(SPInProp). Here we focus on an accelerated solution of pPDEs, but SPInProp can
potentially accelerate various computational tasks in deep learning models.

\section{General approach to SPInProp}

In this section, we set up key definitions and put forth the main approach.  We
first define the LRNR architecture in Section~\ref{sec:lrnr}, and present the
associated FastLRNR architecture in Section~\ref{sec:fastlrnr}.

\subsection{Low Rank Neural Representation (LRNR)} \label{sec:lrnr}

Given $n \in \NN$, denote $[n] := \{1, ... , n\}$ and $[n]_0 = [n] \cup \{0\}$.
A \emph{feedforward NN} for given depth $L \in \NN$ and widths $(M_0, ...\,,M_L)
\in \NN^{L+1}$ is a function $f: \RR^{M_0} \to \RR^{M_L}$ defined as the
sequence of compositions 
\begin{equation} \label{eq:NN}
    \zell = \sigma( \Well \zellm + \bell ),
    \quad
    \ell \in [L-1],
    \quad
    z^L = W^L z^{L-1} + b^L,
\end{equation}
where the weight matrices $\Well \in \RR^{\Mell \times \Mellm}$ and bias vectors
$\bell \in \RR^{\Mell}$ for $\ell \in [L]$ are called the parameters of the NN,
and the nonlinear activation function $\sigma: \RR \to \RR$ acts entrywise.  The
input to the NN is $\zinput \in \RR^{M_0}$ and the output is $\zoutput \in
\RR^{M_L}$. We denote the maximum width by $\Mmax := \max_{\ell \in [L]_0}
M_\ell$, the total width by $\Mtotal := \sum_{\ell \in [L]_0} M_\ell$, the
collection of the weights and biases $W := (\Well)_{\ell \in [L]}$ and $b :=
(\bell)_{\ell \in [L]}$, respectively.

\emph{Implicit Neural Representation} (INR) is a dense NN whose input dimension
$M_0$ is the spatio-temporal dimension and its output dimension $M_L$ is the
vector dimension of physical variables (density, temperature, pressure, etc). So
the dimensions $M_0$  and $M_L$ are relatively small. For example, an INR
representing a solution to a scalar PDE in $\RR^2$ has $M_0 = 2$ and $M_L = 1$.

To define LRNRs, we start by defining an INR $\uinr: \RR^{M_0} \to \RR^{M_L}$
whose weight matrix $W^\ell$ is substituted by the product of three weight
matrices
\begin{equation} \label{eq:svd}
  \Well = \Uell \Sell \Vellt,
  \quad
  \ell \in [L],
\end{equation}
with the individual factors taking on the form $\Uell := [\Uell_1 \mid \cdots
\mid \Uell_\rell] \in \RR^{\Mell \times \rell},$ $\Vell := [\Vell_1 \mid \cdots
\mid \Vell_\rell] \in \RR^{\Mell \times \rell},$ $\Sell := \diag (\sell) \in
\RR^{\rell \times \rell}.$ We denote $\sell = (\sell_1, \cdots , \sell_\rell)
\in \RR^\rell$ and assume all entries in $\sell$ to be non-negative.  We let $r
:= (\rell)_{\ell \in [L]}$, $\rmax := \max_{\ell} r_{\ell \in [L]}$, $\rmin :=
\min_{\ell \in [L]} r_\ell$ and $\rtotal := \sum_{\ell \in [L]} \rell$. 

The factored re-formulation \eqref{eq:svd} of the weight matrices \eqref{eq:NN}
resembles the singular value decomposition (SVD) in its appearance
\cite{golub96}, and if we assume $\rell \ll \Mell$ for all $\ell \in [L]$ the
products \eqref{eq:svd} form low rank matrices.  Note that the same treatment is
possible for the bias terms $\bell$, but we will not factor the bias here
(however, see \cite{rim2024} where the bias is similarly factored and plays a
central role).

We collect these matrices with the notations $U := (\Uell)_{\ell \in [L]},$ $V
:= (\Vell)_{\ell \in [L]},$ and $s := (\sell)_{\ell \in [L]}$.  We refer
specifically to parameters in $s$ as \emph{coefficient parameters}, and to the
parameters in $(U, V)$ as \emph{bases parameters}; they will play distinct
roles. Note that $s$ has $\cO(L\rmax)$ parameters, whereas $U$ and $V$ each
have $\Omega(\Mmax \rmin)$ parameters. The quadruple $(U, V, b, s)$ contains
all the parameters of $\uinr$ so one can express its full parametric dependence
by writing
\begin{equation} \label{eq:inr}
  \uinr(\bigcdot) = \uinr(\bigcdot \,; U, V, b, s).
\end{equation}

The meta-learning framework proposed in \cite{cho2023} has two different phases
for training the parameters $(U, V, b, s)$. In the \emph{first phase}, or the
\emph{meta-learning phase}, the bases parameters and the biases $(U, V, b)$ are
learned. Once learned, they are fixed as $(\Ufix, \Vfix, \bfix)$, leaving us
with a famliy of INRs
\begin{equation} \label{eq:lrnr}
  \ulrnr(\bigcdot\,; s) := \uinr(\bigcdot\,; \Ufix, \Vfix, \bfix, s),
  \quad
  s \in \RR^{\rtotal}.
\end{equation}
If we suppose $\rell \ll \Mmax$ then we have $\rtotal \ll \Mtotal$, and $\ulrnr$
depends now solely on a small number of coefficient parameters in $s$, although
it can generally have large widths.  We refer to this low-dimensional family of
INRs as a \emph{Low Rank Neural Representation} (LRNR).  Upon the completion of
the meta-learning phase only the coefficient parameters $s$ are assumed to be
trainable, and they are trained during the \textit{second phase}, or the
\emph{fine-tuning phase}.  We defer further details to
Section~\ref{sec:fastlrpinns}.

We now discuss the meta-learning framework in the context of solving pPDEs.
Many scientific computing problems are naturally posed as pPDE
problems~\cite{hesthaven2016}, and in a pPDE, there is a vector of relevant
physical parameters $\mu$ lying in a domain $\cD \subset \RR^p$ $(p \in \NN)$.
The goals of the two phases are, respectively: (1) Find a LRNR with trained
bases and biases tailored to the pPDE; (2) For queried $\mu$, determine the
coefficient parameters $s(\mu)$ that yields the approximation of the pPDE
solution $u(\bigcdot\,; \mu)$ in the LRNR form
\begin{equation} \label{eq:lrnr_solved}
  u(\bigcdot\,; \mu)
  \approx
  \ulrnr (\bigcdot\,; s(\mu)) 
  =
  \uinr (\bigcdot \,; \Ufix, \Vfix, \bfix, s(\mu)).
\end{equation}
The relationship $s(\mu)$ is learned by backpropagation-based approaches, using
$\ulrnr$ as an ansatz in the pPDE. This meta-learning framework was named the
Low Rank PINNs (LR-PINNs\footnote{LRNR refers specifically to the architecture,
and LR-PINNs to the overarching meta-learning framework using LRNRs.}) in
\cite{cho2023}.

\subsection{FastLRNRs} \label{sec:fastlrnr}

We turn our attention to the dimensionality reduction of LRNRs.  The key idea is
to construct an approximation to the $s$-independent parts of $\ulrnr$
\eqref{eq:lrnr}.  Taking two contiguous layers, we write the part between the
coefficient parameters $s^\ell$'s as $\rho^{\ell}(\bigcdot) := \Vellpt
\sigma(\Uell \bigcdot)$ and view it as a projected version of the nonlinear
activation $\sigma$ (we omit the biases for ease of exposition).  It is known
that when a scalar nonlinear function is applied to low rank vectors, the
function values themselves can be well-approximated by low rank bases in certain
situations \cite{everson95,barrault04,chanturantabut10,carlberg2013, %
hesthaven2016}.  A family of well-known and simple techniques can be applied to
obtain an approximation to $\rho^\ell$ of the form
\begin{equation} \label{eq:zeta}
  \zetaell (\bigcdot)
  :=
  \hVellpt \sigma( \hUell \,\bigcdot ),
  \quad
  \ell \in [L-1],
\end{equation}
in which the reduced bases parameters now have the dimensions $\hUell \in
\RR^{\hrell \times \rell}$ and $\hVellp \in \RR^{\hrell \times \rellp}$, and the
reduced dimensions are small in the sense $\hrell \sim \rell \ll \Mmax$.  We write
$\hat{r} := (\hrell)_{\ell \in [L-1]}$ and $\hrellmax := \max_{\ell \in [L-1]}
\hrell$.

For example, consider the use of the empirical interpolation method (EIM/DEIM
\cite{barrault04,chanturantabut10}). Writing out the EIM approximation, which
uses a subsampling of the state vectors, we have
\begin{equation} \label{eq:eim}
  \rho^\ell(\bigcdot)
  \approx
  \zetaelleim(\bigcdot)
  :=
  \Vellpt
  \Xiell (\Pellt \Xiell)^{-1}
  \sigma( \Pellt \Uell \, \bigcdot ),
  \qquad
  \Xiell \in \RR^{\Mell \times \hrell},
  \quad
  \hrell \in \NN,
\end{equation}
where $\Pell \in \RR^{\Mell \times \hrell}$ is a sampling matrix made up of
subcolumns of a $\Mell \times \Mell$ identity matrix, and $\Xiell \in
\RR^{\Mell \times \hrell}$ is a matrix with linearly independent columns whose
square subblock $\Pellt \Xiell \in \RR^{\hrell \times \hrell}$ is invertible.
Viewed in the general form \eqref{eq:zeta}, this approximation has $\hUell =
\Pellt \Uell$ and $\hVellp =  \Vellpt \Xiell (\Pellt \Xiell)^{-1}.$

Inserting \eqref{eq:zeta} into the LRNR architecture, one obtains $\hzell =
\zetaell( \diag(\sell ) \hzellm)$ where the multiplication by $\diag(s^\ell )$
can be rewritten using the notation $\odot$ for the Hadamard (elementwise)
product. Concisely,
\begin{equation} \label{eq:reduced_LRNR}
  \hzell
  =
  \zetaell( \sell \odot \hzellm),
  \quad
  \ell \in [L-1],
  \quad
  \hat{z}^L
  =
  U^L (s^L \odot \hat{z}^{L-1}).
\end{equation}
Then we have an approximation of the hidden states $\zell$ in the original LRNR
\eqref{eq:NN} purely in terms of the approximate projected/reduced states, that
is 
\[
  \hzell \approx \Vellpt \zell, 
  \quad
  \ell \in [L-1]_0,
  \quad
  \hat{z}^L \approx z^L.
\]
Note that in the input and output layers, the reduction can be trivial.

\begin{figure}
  \centering
  \includegraphics[width=0.99\textwidth]{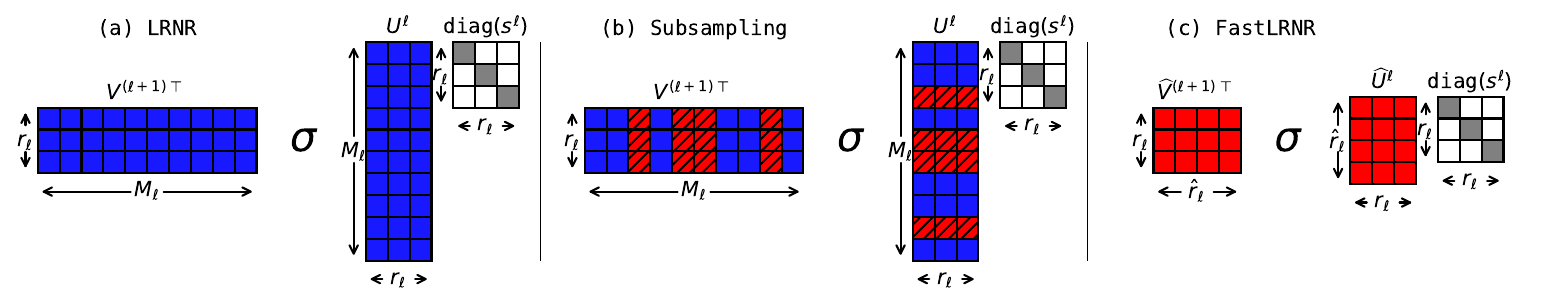}
  \caption{An illustration of the (a) LRNR and the associated (c) FastLRNR
  architectures, obtained by e.g. subsampling the rows of the low rank matrices
  as shown in (b).}
  \label{fig:FastLRNR}
\end{figure}

We refer to the reduced form of LRNR \eqref{eq:reduced_LRNR}, as a
\emph{FastLRNR} $\ufast: \RR^{M_0} \to \RR^{M_L}$. Diagrams in 
Figure~\ref{fig:FastLRNR} compare the LRNR versus the FastLRNR architectures,
and illustrate the subsampling strategy devised in EIM in this context.  We
remark that the approximation still has the structure of NNs \eqref{eq:NN},
except that the nonlinear activation functions $\sigma$ correspond to
specialized functions $\zetaell$, which are generally different functions
depending on the layer (cf. \cite{liu2024}). In this analogy, the Hadamard
products with $\sell$ in \eqref{eq:reduced_LRNR} play the role of affine
mappings in \eqref{eq:NN}, and in this sense FastLRNRs \eqref{eq:reduced_LRNR}
(and LRNRs) are said to be \emph{diagonalizeable}.

\subsection{SPInProp via backpropagation of FastLRNR} \label{sec:spinprop}

SPInProp complexity depends only on $L$ and the small dimension $\rmax$.
Comparing the backpropagation complexities: (1) For NNs \eqref{eq:NN} with
parameters $(W, b)$, $\cO(L \Mmaxz)$ operations; (2) For LRNR $\ulrnr$
\eqref{eq:lrnr} with parameters $s$, $\cO(L \Mmax \rmax)$; (3) For FastLRNR
$\ufast$ \eqref{eq:reduced_LRNR} with parameters $s$, $\cO(L \rmaxz )$.

\section{Fast Low Rank PINNs (Fast-LR-PINNs)} \label{sec:fastlrpinns}

We seek to demonstrate that SPInProp operations can be used to efficiently
perform backpropagation operations when solving pPDEs. To this end, we devise
computational experiments in the LR-PINNs meta-learning framework \cite{cho2023}
with important additions. We discuss the meta-learning phase in
Section~\ref{sec:metalearning}.  A new phase using SPInProp, which we call the
\emph{fast phase}, is introduced in Section~\ref{sec:fastphase}

\subsection{Meta-Learning with Sparsity Promoting Regularization}
\label{sec:metalearning}

As discussed in Section~\ref{sec:lrnr}, during the meta-learning phase the bases
and bias parameters $(U, V, b)$ are trained. This is done using a hypernetwork
representation of $s(\mu)$ (see \eqref{eq:lrnr_solved}). We define a
hypernetwork $\fhyper : \cD \to \RR_+^{\rtotal}$ which is itself a NN
\eqref{eq:NN} whose widths satisfy $\Mhyperell \sim \rtotal$ and has an extra
ReLU applied so that the outputs are non-negative.  We call its NN parameters 
$\Theta$ and write $f(\bigcdot) = f(\bigcdot \,; \Theta)$.

The meta-network is used during this phase, defined as
\begin{equation} \label{eq:umeta}
  \umeta(\bigcdot\, ; \mu, U, V, b, \Theta)
  :=
  \uinr (\bigcdot \,; U, V, b, \fhyper(\mu \,; \Theta)).
\end{equation}
This model results from a type of output-to-parameter concatenation between 
(a) $\uinr$ with
low rank structure in its parameters \eqref{eq:inr}, and (b) 
$\fhyper$ whose non-negative outputs feed into $\uinr$ as coefficients $s$.

The PINNs loss for the meta-learning phase is given by
\begin{equation}
  \begin{aligned}
  \lossmeta(U, V, b, \Theta)
  :&=
  \mathbb{E}_{(\bigcdot\,, \mu)}
  \left[
  \abs{ \cN[\umeta\,; \mu](\bigcdot\, ; \mu, U, V, b, \Theta)}^2
  \right]
  \\
  &+
  \mathbb{E}_{(\bigcdot\,,\mu)}
  \left[
    \abs{ \mathcal{B}[\umeta\,; \mu] (\bigcdot \,; \mu, U, V, b, \Theta) }^2
  \right]
  \end{aligned}
\end{equation}
where $\cN[ \bigcdot\,; \mu]$ denotes a (nonlinear) differential operator, and
$\cB[\bigcdot\,; \mu]$ the initial-boundary operator. The expectation is 
taken over some probability measure over the PDE domain $\Dom$ and physical
parameter domain $\cD$ as specified in the pPDE problem. Uniform
measure is commonly used.

We define the orthogonality regularization term and the sparsity-promoting
regularization term as
\begin{equation}
  \begin{aligned}
  \regortho(U, V)
  &:=
  \sum_{\ell \in [L]} \Norm{\Uellt \Uell - I^\ell}{F}^2
                    + \Norm{\Vellt \Vell - I^\ell}{F}^2,
  \\
  \regsparse(\Theta)
  &:=
  \EE_{\mu}
  \left[
    \sum_{\ell \in [L]}
    \Norm{ \sigma(\Gamma^\ell \fhyper^\ell(\mu; \Theta))}{1}
  \right],
  \end{aligned}
\end{equation}
where $I^\ell \in \RR^{\rell \times \rell}$ are identity matrices,
$\Norm{\bigcdot}{F}$ denotes the Frobenius norm, and where the matrices
$\Gamma^\ell \in \RR^{(\rell-1) \times \rell}$ are banded matrices with -1's on
the diagonal and $\gamma$ on the 1st super-diagonal ($\gamma \ge 1$ is a
hyperparameter). The motivation for adding $\regsparse$ is to promote the
sparsity in the coefficient parameters $s$ in $\uinr$ above \eqref{eq:umeta}, 
thereby constraining the meta-network $\umeta$ to become as low rank as
possible. In $\umeta$ the coefficient parameters are coupled to the $\fhyper$
output variables, so we enforce the sparsity on the parameters $\Theta$ of
$\fhyper$. When $\regsparse$ is zero the coefficient parameters $s^\ell$ (which
are output from $\fhyper^\ell$) satisfy $(1/\gamma) s^\ell_{i} > s^\ell_{i+1}$
so a geometric rate of decay is achieved. An apparently more intuitive
\cite{FoucartRauhut2013} choice of taking the 1-norm $\Norm{\fhyper^\ell(\mu;
\Theta)}{1}$ was less successful in promoting sparsity in our experiments.

Finally, the meta-learning problem is given by
\begin{equation} \label{eq:meta-loss}
  \min_{U, V, b, \Theta}
  \quad
  \lossmeta (U, V, b, \Theta)
  +
  \regparamortho \regortho (U, V)
  +
  \regparamsparse \regsparse (\Theta).
\end{equation}

\subsection{Fast phase} \label{sec:fastphase}

At the end of the meta-learning phase, we have learned the bases and bias
parameters $(\Ufix, \Vfix, \bfix)$, along with the hypernetwork parameters
$\Thetafix$. This brings us to the fine-tuning phase, where one solves the PDE 
for any given physical parameter $\mu \in \cD$ by fine-tuning the $\ulrnr$
coefficient parameters $s$ \eqref{eq:lrnr} as informed by the differential
equations. The coefficients $s$ are initialized at the hypernetwork prediction
value $\fhyper(\mu; \Thetafix)$. During this phase, the learning problem is to
minimize the physics-informed loss function over a small number of coefficient
parameters $s$, 
\begin{equation} \label{eq:optim_tune}
  \begin{aligned}
    &\min_{s} \losstune(s; \mu),
    \\
    &\where
    \losstune(s; \mu)
    :=
    \mathbb{E}
    \left[
    \abs{ \cN[\ulrnr\,; \mu](\bigcdot\,; s)}^2
    \right]
    +
    \mathbb{E}
    \left[
      \abs{ \cB[\ulrnr\,; \mu ](\bigcdot\, ; s)}^2
    \right].
  \end{aligned}
\end{equation}

In the previous experiments \cite{cho2023}, the fine-tuning phase for $\ulrnr$
was found to reliably yield improved PDE solutions. Now, if FastLRNR $\ufast$
\eqref{eq:reduced_LRNR} is a type of an approximation of $\ulrnr$, it is natural
to attempt to update the coefficient parameters $s$ using the smaller $\ufast$,
and test if these updates result in improved solutions. If they do, it means
updates using $\ufast$ alone can lead to improved coefficient values 
$s_{\textrm{fast}}(\mu)$, at the computationally cheaper cost of SPInProp
operations (Section~\ref{sec:spinprop}).

We newly introduce the fast phase during which we solve a learning problem for
$\ufast$. Following a subsampling strategy (Section~\ref{sec:fastlrnr}),
we define a sparsely sampled version of the fine-tuning-phase loss $\losstune$,
\begin{equation} \label{eq:lossfast}
  \begin{aligned}
    \lossfast(s; \mu)
    :=&
    \,
    \sum_{\bigcdot\,\, \in \cX \cap \Dom}
    \abs{ \cN[\ufast\,; \mu](\bigcdot \,; s)}
    +
    \sum_{\bigcdot\,\, \in \cX \cap \partial \Dom}
      \abs{ \cB[\ufast\,; \mu ](\bigcdot\, ; s) },
    \end{aligned}
\end{equation}
where $\cX \subset \Dom \cap \partial \Dom$ is a set of sparse sampling points.
Here we have used the $\ell_1$ error, motivated by the $L_1$ minimization
formulation of convective problems \cite{guermond2007,guermond2008}.

A na\"ive approach would be to substitute the higher-dimensional learning
problem \eqref{eq:optim_tune} by a lower-dimensional one, replacing the loss
$\losstune$ with a direct analogue $\lossfast$.  However, this approach did not
automatically lead to improvements in the solution in our experiments.  While
the training loss decreases consistently, the FastLRNR solution
$\ulrnr(\bigcdot\,; s)$ eventually diverges away from the true solution despite
improvement during the initial epochs.  This suggests the na\"ive learning
problem is affected by a type of generalization issue.  We speculate there are
two possible causes.  First, FastLRNR performs sparse sampling of both the PDE
domain and the hidden states, possibly incurring bias that leads to
generalization errors.  Second, FastLRNR performs projections at each layer and
the effect of these projections on stability can be significant and complex,
potentially causing this issue; related stability issues arise in simpler linear
models like reduced basis methods (see e.g.
\cite{rozza2007,gener2011,gerner2012,dahmen2014}).

To deal with the generalization issue, we regularize the na\"ive learning
problem by adding a locality regularization term for the coefficient parameters
$s$. We set the fast phase learning problem as
\begin{equation}\label{eq:optim_fast}
  \min_{s} \,\,
    \lossfast (s; \mu)
    +
    \regparamlocal \reglocal (s; \mu),
  \quad
  \reglocal (s; \mu)
  :=
  \sum_{\ell \in [L]}\Norm{s^\ell - \fhyper^\ell(\mu)}{1}.
\end{equation}
Once this optimization problem is solved, the solution $s_{\textrm{fast}}(\mu)$
can be used in the original LRNR representation.  Note that the problem
\eqref{eq:optim_fast} was informed by physics at the sampling points $\cX$,
implying that $\ufast$ is trained to be close to $\ulrnr$ only at these points.
To obtain a solution that is accurate for the entire domain $\Dom$, we make use
of the original LRNR architecture, setting as the solution
$\ufastsol(\bigcdot\,;\mu) := \ulrnr (\bigcdot \,; s_{\textrm{fast}}(\mu)).$ In
short, we return to the achitecture with possibly large widths, once the
coefficient parameters $s$ are updated via gradient descent using only $\ufast$
and SPInProps.

\begin{figure}
  \centering
  \begin{tabular}{c}
  \includegraphics[height=0.16\textheight]{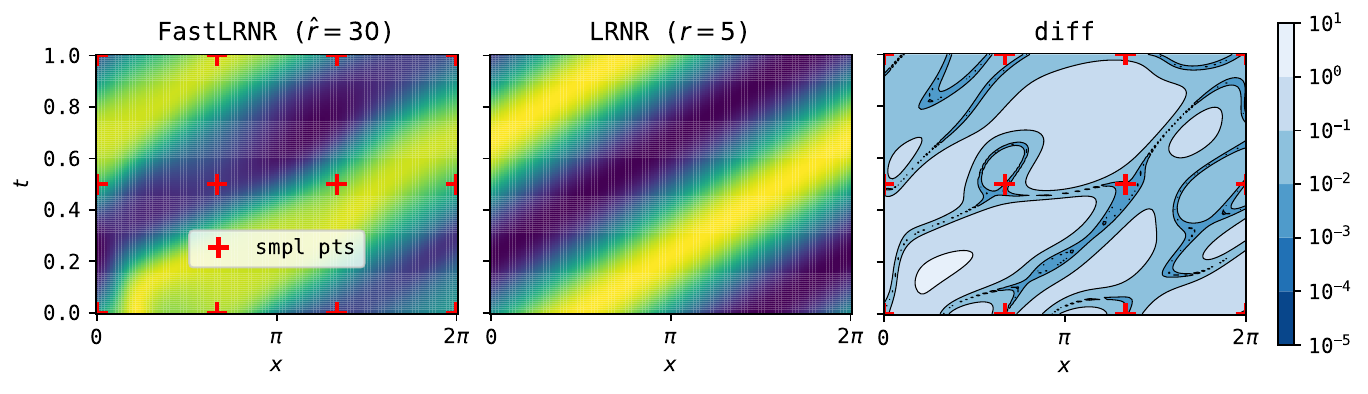}
  \\
  \includegraphics[height=0.16\textheight]
    {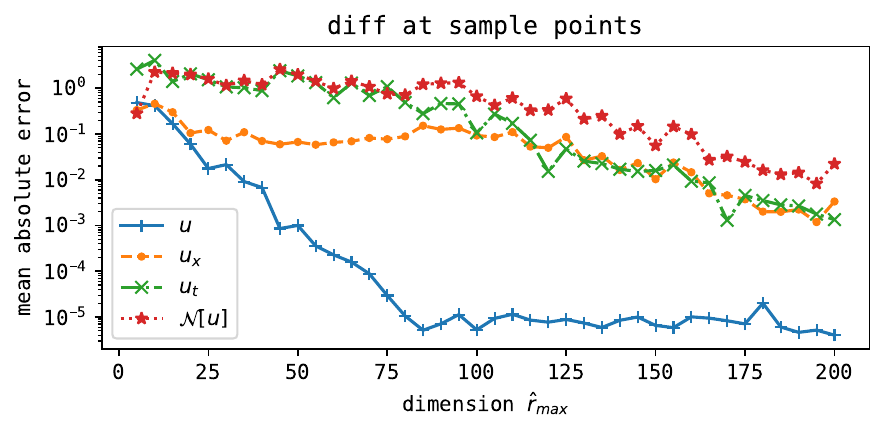}
  \end{tabular}
  \caption{The pointwise value difference between the FastLRNR $\ufast$ and LRNR
  $\ulrnr$ for the convective case $\domone$ \ref{eq:param_dom} with physical
  parameter $\mu = (7, 0, 0)$ and dimension $\hrellmax = 30$. The sparse
  sampling points in $\cX$ are marked by {\tt +}. The FastLRNR is a much smaller
  than LRNR but agrees with it at the sampling points (top).  SPInProp
  accuracy versus discrete dimension $\hrellmax$ ($\ell = 2,3,4$), mean absolute
  error over the uniform sampling points $\cX \subset \Omega$ (bottom).}
  \label{fig:eim_pts}
\end{figure}
  
\section{Parametrized convection-diffusion-reaction problem}
\label{sec:ppde}

We consider a parametrized initial boundary value problem: 
Convection-diffusion-reaction on the spatio-temporal domain $\Omega := (0, 2\pi)
\times (0, 1)$ with periodic boundary conditions. We seek the solution $u: \Dom
\to \RR$ satisfying the following PDE with non-negative physical parameters $\mu
= (\mu_1, \mu_2, \mu_3)$,
\begin{equation} \label{eq:pde}
  \begin{aligned}
    \cN[u\,; \mu]
    &=0,
    \quad
    (x, t) \in \Omega,
    \\
    & \quad \where
    \cN [u\,; \mu] 
    := 
    \partial_t u + \mu_1 \partial_x u - \mu_2 \partial_{xx} u - \mu_3 u (1 - u),
    \\
    u(x, 0)
    &=
    \sin(x),
    \quad 
    x \in (0, 2\pi),
    \quad
    u(0, t)
    =
    u(2\pi, t),
    \quad
    t \in (0, 1).
  \end{aligned}
\end{equation}
We consider two physical parameter domains: The first is a pure convection
problem, and the second is the full convection-diffusion-reaction problem: 
\begin{equation} \label{eq:param_dom}
  \domone
  =
  \diag([\muimin, \muimax]),
  \quad
  \domtwo
  =
  \diag([\muiimin, \muiimax]),
\end{equation}
where we set $\muimin = (5, 0, 0)$, $\muimax = (8, 0, 0)$, $\muiimin = (1, 0,
0)$, $\muiimax = (3, 2, 2)$. The PDE is nonlinear and is highly convective when
the diffusion term $\mu_2$ is small. This is a numerical example from
\cite{cho2023} where the PDE was used in extreme regimes to test the
effectiveness of the meta-learning approach.

Our PyTorch-based \cite{PGF+19} implementation of FastLRNR is based on that of
\cite{cho2023}, and is available in a public code repository
\cite{FastLRNRRepo}. We refer the readers to the repository for the precise
details that are omitted here.

Upon completing the meta-learning phase using a meta-network of size $\Mmax =
4000$ and $\rmax = 100$, most of the predicted coefficient outputs from
$\fhyper$ are identically zero. These coefficients are dropped and the
corresponding columns of bases parameters $(U, V)$ are truncated. The
coefficient dimensions of LRNRs are adjusted correspondingly, yielding $r = (1,
5, 5, 3, 1)$ for $\domone$ and $r = (1, 49, 30, 25, 1)$ for $\domtwo$. This
computation implies that LRNR solutions with $\rmax \ll \Mmax$ exist for this
pPDE for $\domone$ and $\domtwo$.

Next, we construct the associated FastLRNR \eqref{eq:reduced_LRNR}. We proceed
by evaluating $\ulrnr$ at fixed input sampling points $\cX$ and their
perturbations, across various parameters values in $\cD$. We choose 4-by-3
uniform grid points as the $12$ input sampling points $\cX$, then we use EIM to
compute $(\Xi^\ell, P^\ell)$ \eqref{eq:eim}.  Since $\ufast$ and $\ulrnr$ share
the coefficient parameters, comparing them as spatio-temporal functions on
$\Omega$ can be instructive. In Figure~\ref{fig:eim_pts}, $\ufast$ with
$\hrellmax = 30$ is compared with $\ulrnr$, with coefficients initialized as
$\fhyper(\mu)$ and $\mu = (7, 0, 0)$ for both. We see that they (and their
derivatives) agree at the sampling points.

\begin{figure}
  \centering
  \includegraphics[width=0.95\textwidth]{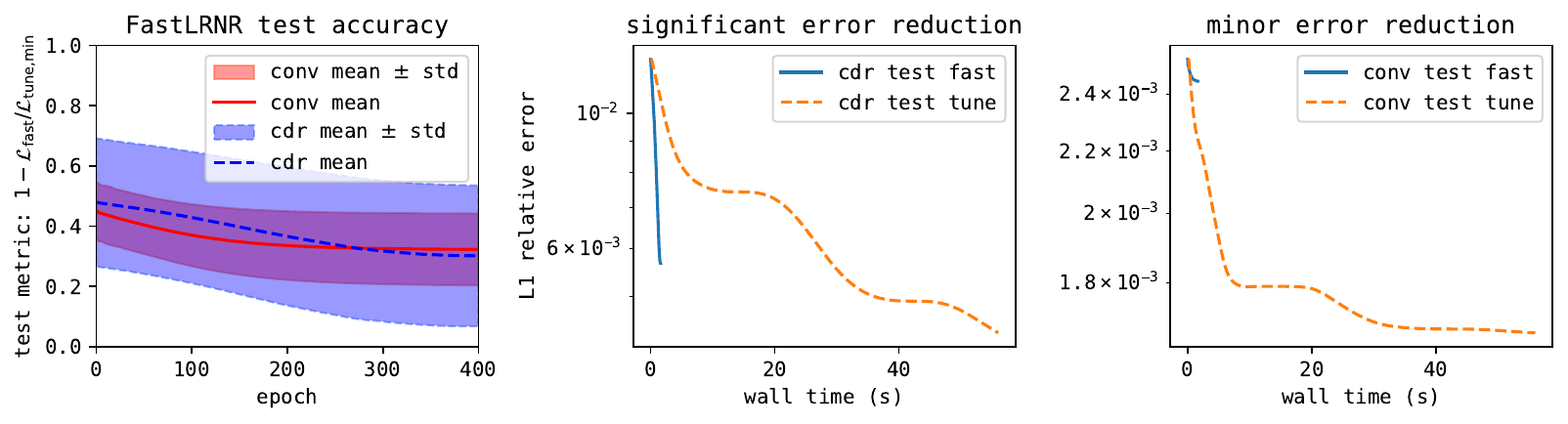}
  \caption{Plots comparing the accuracy of FastLRNR and LRNR over 20 test cases.
  Comparing the FastLRNR test loss to the minimal test loss $\losstune$ achieved
  with LRNR, FastLRNR cheaply obtains nearly half of the test loss accuracy of
  LRNR (left). To the right are shown two $L^1$ relative error plots versus
  wall time. In the majority (16 out of 20) of test examples, within one second
  of starting the optimization problem \eqref{eq:optim_fast}, FastLRNR achieves
  slightly worse $L^1$ relative error compared to LRNR (middle). In a minority
  of test examples (4 out of 20), only a minor error reduction is achieved,
  although in these cases errors were lower overall (right).}
  \label{fig:test_accuracy}
\end{figure}

Finally, the learning problems for the FastLRNR \eqref{eq:optim_fast} and the
LRNR \eqref{eq:optim_tune} are solved using Adam \cite{KingBa15} up to 400
epochs for 20 test examples (10 in $\domone$ and 10 in $\domtwo$).  The test
accuracy of these solutions are shown in Figure~\ref{fig:test_accuracy}.  In
most examples, FastLRNR achieves $L^1$ relative error close to that of LRNR for
$\hrellmax = 5$, and this accuracy is achieved within a fraction of a second, 
owing to the computational speed of SPInProp operations
(Section~\ref{sec:spinprop}).  In some examples, the FastLRNR does not improve
in accuracy, however in these subcases we found that the initial guess for the
coefficients from $\fhyper$ was already very accurate.

The FastLRNR is roughly $\times 36$ faster than fine-tuning-phase training (for
$\Mmax = 4000$, $\rmax=49$, $\hrellmax=5$).  Wall time for a single Adam step
using SPInProp is 0.004s versus 0.14s for standard backpropagation on an NVIDIA
V100 GPU with 32GB memory. In principle the speedup would be greater for even
larger widths, since the complexity scales linearly with respect to the widths.
Moreover, the fast phase can run on the CPU due to its small memory
requirements.

\section*{Acknowledgements}
The authors acknowledge the Research Infrastructure Services (RIS) group at
Washington University in St.  Louis for providing computational resources and
services needed to generate the research results delivered within this paper
(URL {\tt ris.wustl.edu}).  K. Lee acknowledges support from the U.S. National
Science Foundation under grant IIS 2338909.

\bibliographystyle{siamplain}

\end{document}